\documentclass[lettersize,journal]{IEEEtran}
\usepackage{amsmath,amsfonts}
\usepackage{algorithmic}
\usepackage{algorithm}
\usepackage{array}
\usepackage[caption=false,font=normalsize,labelfont=sf,textfont=sf]{subfig}
\usepackage{textcomp}
\usepackage{stfloats}
\usepackage{url}
\usepackage{verbatim}
\usepackage{graphicx}
\usepackage{cite}
\usepackage{colortbl}
\usepackage{multirow}
\usepackage{graphicx}
\usepackage{amsthm}
\usepackage{mathdesign}
\usepackage{xcolor}
\usepackage{amsfonts}
\usepackage{makecell}
\usepackage{booktabs}
\usepackage{bbding}

\usepackage{float}
\usepackage{xspace}
\newcommand{\ie}{{\emph{i.e.}}\xspace}
\newcommand{\eg}{{\emph{e.g.}}\xspace}

\usepackage[switch]{lineno}
\usepackage{amssymb}
\usepackage{pifont}
\newcommand{\cmark}{\ding{51}\xspace}%
\newcommand{\xmarkg}{\textcolor{lightgray}{\ding{55}}\xspace}%

\begin{document}
\title{Open-set Anomaly Segmentation in Complex Scenarios}

\author{
  Song Xia, Yi Yu, Henghui Ding, Wenhan Yang, \textit{Member, IEEE}, Shifei Liu,\\ Alex C. Kot, \textit{Life Fellow, IEEE}, Xudong Jiang, \textit{Fellow, IEEE}
  
\IEEEcompsocitemizethanks{
        \IEEEcompsocthanksitem Song Xia, Shifei Liu, Alex C. Kot, and Xudong Jiang are with the Rapid-Rich Object Search Lab, School of Electrical and Electronic Engineering, Nanyang Technological University, Singapore, (e-mail: \{xias0002, sliu052, eackot, exdjiang\}@ntu.edu.sg).
        \IEEEcompsocthanksitem Yi Yu is with the Rapid-Rich Object Search Lab, Interdisciplinary Graduate Programme, Nanyang Technological University, Singapore, (e-mail: yuyi0010@e.ntu.edu.sg).
        \IEEEcompsocthanksitem Wenhan Yang is with Pengcheng Laboratory, Shenzhen, China, (e-mail:  yangwh@pcl.ac.cn).
        \IEEEcompsocthanksitem Henghui Ding is with the Institute of Big Data, Fudan University, Shanghai, China, (e-mail:  henghui.ding@gmail.com).
	}
}


\maketitle
\begin{figure*}
\begin{center}
    \includegraphics[width=1\textwidth]{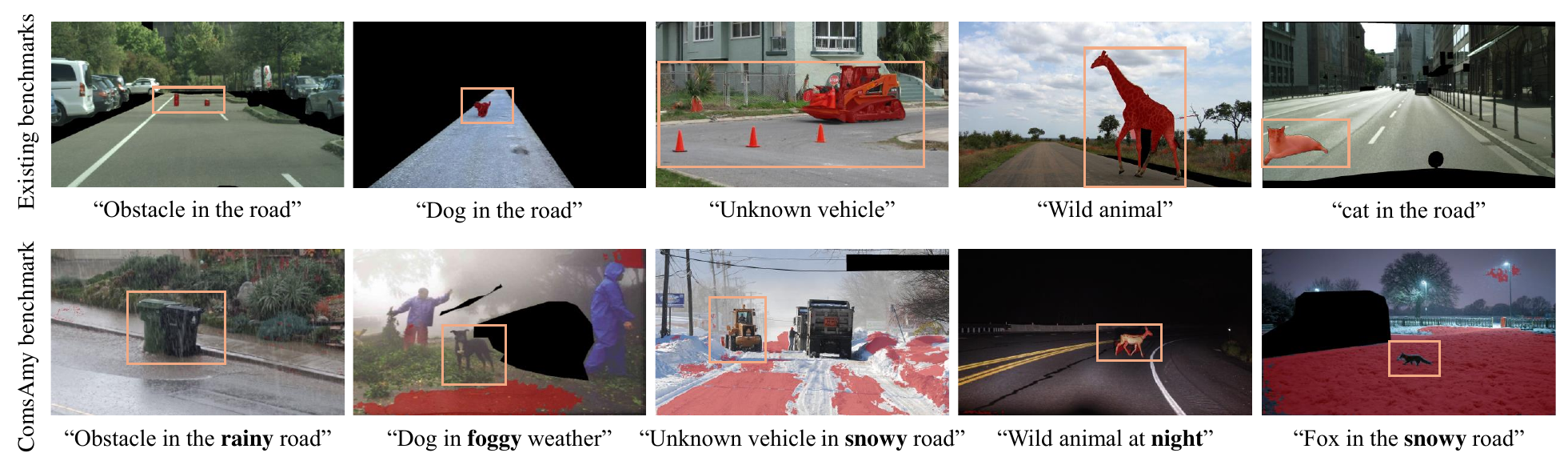}
    \caption{Visualization of anomalous segmentation results by SoTA Mask2Anomaly~\cite{rai2024mask2anomaly} model on existing and our benchmarks. 
    We highlight the anomalies using yellow bounding boxes.
    The pure black regions denote void pixels, which are excluded from evaluation.
    The top row presents anomalous images from existing benchmarks, including FS-L\&F~\cite{blum2019fishyscapes}, SMIYC-Obstacle~\cite{chan2021segmentmeifyoucan}, RoadAnomaly~\cite{lis2019detecting}, SMIYC-Anomaly~\cite{chan2021segmentmeifyoucan}, and FS-Static~\cite{blum2019fishyscapes}.
    The bottom row displays those from ours. 
    Objects marked with red color are predicted as anomalies by the Mask2Former model.
    The comparative analysis reveals that the model exhibits significant limitations under challenging conditions, such as rainy weather and low illumination.
} 
    \label{fig:visualized result}
\end{center}
\end{figure*}

\begin{abstract}
Precise segmentation of out-of-distribution (OoD) objects, herein referred to as anomalies, is crucial for the reliable deployment of semantic segmentation models in open-set, safety-critical applications, such as autonomous driving.
Current anomalous segmentation benchmarks predominantly focus on favorable weather conditions, resulting in untrustworthy evaluations that overlook the risks posed by diverse meteorological conditions in open-set environments, such as low illumination, dense fog, and heavy rain.
%
%
To bridge this gap, this paper introduces the ComsAmy, a challenging benchmark specifically designed for open-set anomaly segmentation in complex scenarios.
ComsAmy encompasses a wide spectrum of adverse weather conditions, dynamic driving environments, and diverse anomaly type to comprehensively evaluate the model performance in realistic open-world scenarios.
Our extensive evaluation of several state-of-the-art anomalous segmentation models reveals that existing methods demonstrate significant deficiencies in such challenging scenarios, highlighting their serious safety risks for real-world deployment.
To solve that, we propose a novel energy-entropy learning (EEL) strategy that integrates the complementary information from energy and entropy to bolster the robustness of anomaly segmentation under complex open-world environments.
Additionally, a diffusion-based anomalous training data synthesizer is proposed to generate diverse and high-quality anomalous images to enhance the existing copy-paste training data synthesizer
%
Extensive experimental results on both public and ComsAmy benchmarks demonstrate that our proposed diffusion-based synthesizer with energy and entropy learning (DiffEEL) serves as an effective and generalizable plug-and-play method to enhance existing models, yielding an average improvement of around 4.96\% in $\rm{AUPRC}$ and 9.87\% in $\rm{FPR}_{95}$.
\end{abstract}


\section{Introduction}
\IEEEPARstart{R}ecent spectacular advances in neural networks steadily expand the horizon of exciting applications of computer vision~\cite{kirillov2023segment,rombach2022high,huang2022pixel,luo2023beyond,wang2025segment}. 
For many downstream computer vision applications, the complex image understanding functionality provided by the advanced semantic segmentation model is crucial~\cite{farabet2012learning}. 
However, while achieving precise pixel-wise classification on inlier classes, those models easily misclassify the anomalous objects, \ie, instances outside the distribution of the training data, into the known classes with a high confidence.
This inability to recognize anomalies greatly hinders its application in real-world safety-critical scenarios like autonomous driving. 
In a semantic segmentation-based self-driving system controlling a car on the road, misclassifying anomalies (such as unexpected animals) could lead to fatal car accidents.
Several anomalous segmentation benchmarks, including Fishyscapes~\cite{blum2019fishyscapes}, SMIYC~\cite{chan2021segmentmeifyoucan}, LostandFound~\cite{pinggera2016lost}, and RoadAnomaly~\cite{lis2019detecting}, have been proposed to evaluate the performance of semantic segmentation models in detecting anomalous objects.
However, these datasets primarily focus on models' ability to distinguish anomalous objects under clear weather conditions, neglecting the complex and dynamic weather conditions characterized by real-world scenarios, such as rain and low illumination. 
\textbf{Consequently, even models renowned for their robust performance in existing benchmarks may experience significant degradation under adverse weather conditions~\cite{rai2024mask2anomaly}.}
This may potentially leading to critical safety hazards while deployed in open-set applications.
As shown in \figurename~\ref{fig:visualized result}, the state-of-the-art (SOTA) Mask2anomaly~\cite{rai2024mask2anomaly} faces a great performance degradation under adverse weather conditions.

Additionally, while copy-paste~\cite{chan2021entropy} synthesized anomaly datasets are leveraged by most SOTA anomaly segmentation methods, this anomalous training data brings two caveats. 
The copy-paste method involves extracting out-of-distribution objects from another closed-set dataset like COCO~\cite{lin2014microsoft} or ADE20K~\cite{zhou2017scene} and pasting them into the training dataset like Cityscapes~\cite{cordts2016cityscapes} to generate the training images. 
%
%
However, the anomalous images generated by the direct copy-paste tend to lack realism, which predisposes the segmentation model to overfitting.
Meanwhile, the anomalous objects sourced from a complementary closed-set dataset exhibit limited diversity, thereby diminishing the model's capacity to generalize to a more complex and challenging scenario.
Consequently, even well-trained models are prone to failure when confronted with challenging and unforeseen anomalies in the open world. 
%

To bridge those two gaps, this paper first introduces the ComsAmy benchmark that focuses on evaluating the anomalous segmentation models under complex and adverse meteorological scenarios in open-set environments.
Furthermore, we propose a diffusion-based anomaly data synthesizer and a novel energy-entropy learning (DiffEEL) strategy to improve model generalization and performance in such intricate open-world scenarios.
%
%
The ComsAmy benchmark supplements existing benchmarks by considering a wide spectrum of adverse weather conditions, varied driving scenarios, and diverse anomaly types, serving as the first anomaly segmentation benchmark to simulate complex open-world conditions.
Our benchmark comprises 468 anomalous images that depict 48 distinct anomaly types across 7 unique landforms.
To ensure that the ComsAmy benchmark effectively captures the challenging conditions encountered under the open-set environment, it encompasses seven representative weather conditions, including rain, night, frost, fog, blizzard, sandstorm, and clear weather, alongside a wide range of driving scenarios, including highways, urban roads, country roads, forests, polar regions, grasslands. and deserts.
Furthermore, the ComsAmy benchmark incorporates the most commonly encountered anomaly types in road scenes, including static and dynamic obstacles, animals, and unidentified or unconventional vehicles.

The proposed diffusion-based anomaly data synthesizer allows the automatic generation of diverse and realistic anomalous training images. 
This enables the generation of anomalous training data with a huge quantity and diversity of anomalous objects without requiring any human labor. 
The synthesizer consists of three key components.
Firstly, a diffusion-based object generator is utilized to generate diverse and natural anomalies, each at a prominent position. 
Secondly, a salient object extractor is employed to extract the precise mask of the generated anomaly. 
Finally, a harmonization blender is utilized to merge anomalous objects, from both diffusion-generated objects and other close-set datasets, into original images.
By taking into account the position, size, and illumination relationships between anomalies and original images, it ensures that the synthesized image is as realistic as possible.
%

Furthermore, we propose a novel energy-entropy learning (EEL) strategy to enhance the anomalous segmentation. 
%
First, the complementary information provided by entropy and energy is utilized to constrain the prediction uncertainty of outliers, thus deriving a more robust anomaly score. 
Additionally, a binary cross-entropy loss is utilized to enlarge the gap of the energy response between the misleading inliers and outliers.
Overall, the main contributions are summarized as follows: 

\begin{itemize}
\item[1.] We introduce ComsAmy, a challenging anomaly segmentation benchmark that encompasses a wide spectrum of adverse weather conditions, dynamic driving environments, and diverse anomaly types to simulate the complex open-set scenarios.
Our evaluation of several SOTA anomalous segmentation models reveals substantial performance degradation under such challenging conditions, indicating serious concerns regarding their safety application in practical open-set scenarios.
\item[2.] We propose a diffusion-based anomaly data synthesizer and a novel energy-entropy learning (DiffEEL) strategy to enhance existing models under complex and diverse open-world scenarios. 
\item[3.] Extensive experiments on both public benchmarks and the ComsAmy benchmark demonstrate the effectiveness of the proposed DiffEEL strategy, which serves as a plug-and-play method that significantly enhances the performance of existing SOTA models.

\end{itemize}


\section{Related Work}

\subsection{Anomaly Segmentation Benchmarks} 
The anomaly segmentation task originated from anomaly classification tasks, where the primary focus is on a model's capability to distinguish images from distributions that differ significantly from the training distributions\cite{meinketowards,lee2018simple,zhang2023global,zhao2017consensus,huyan2022unsupervised,Juimage2015}, such as distributions separating the CIFAR10~\cite{krizhevsky2009learning} from SVHN~\cite{netzer2011reading}.
However, anomaly segmentation demands a more precise, pixel-level identification of anomalous objects within images.

Existing widely used anomaly segmentation benchmarks include the LostAndFound~\cite{pinggera2016lost}, Fishyscapes~\cite{blum2019fishyscapes}, StreetHazards~\cite{basart2022scaling}, Segment-Me-If-You-Can (SMIYC)~\cite{chan2021segmentmeifyoucan}, and RoadAnomaly~\cite{lis2019detecting}. 
The LostAndFound~\cite{pinggera2016lost} considers 9 types of anomalous objects in various street scenes in Germany.
However, this benchmark primarily emphasizes urban environments similar to Cityscapes~\cite{cordts2016cityscapes}, and includes anomalies such as children and bicycles that are part of the Cityscapes training set.
Later, Fishyscapes~\cite{blum2019fishyscapes} filters and refines the data in LostAndFound and proposes Fishyscapes L\&F.
Nevertheless, both benchmarks exhibit limited diversity in terms of anomalies and scenarios, restricting their ability to provide comprehensive evaluations for open-set applications.
The StreetHazards~\cite{basart2022scaling} and Fishyscapes Static~\cite{blum2019fishyscapes} benchmarks increase anomaly diversity through the introduction of synthetic objects.
However, this reliance on synthetic anomalies creates a notable gap between evaluation scenarios and real-world applications.
The Segment-Me-If-You-Can (SMIYC)~\cite{chan2021segmentmeifyoucan}, and RoadAnomaly~\cite{lis2019detecting} benchmarks offer a wide array of anomalies and road-related obstacles. 
However, these benchmarks do not adequately account for diverse weather conditions and complex driving scenarios commonly encountered in open-world environments. 

As a result, models that perform well under existing benchmarks frequently exhibit diminished performance when confronted with challenging real-world circumstances, such as low illumination and dense fog.
Consequently, this leads to a critical gap in evaluating anomaly segmentation models under an intricate and diverse open-world environment.
Motivated by these limitations, we propose the ComsAmy benchmark—the first benchmark explicitly designed to incorporate diverse weather conditions, intricate driving scenarios, and a broad variety of anomalies, providing a comprehensive framework for evaluating segmentation performance in complex open-world environments.

%

\subsection{Anomaly Segmentation Methods}
Previous methods have shown their effectiveness in detecting image~\cite{liang2017enhancing,tian2021weakly,hein2019relu,hendrycks2018deep,meinketowards} or video~\cite{yu2023video,cong2013video} level OoD inputs, either based on uncertainty estimation or modifying the training procedure. However, most methods are not suitable for detecting pixel-wise anomalies for semantic segmentation. 

\textbf{Uncertainty and reconstruction-based anomaly segmentation.} 
Inspired by the intuition that the anomaly should get a higher uncertain prediction result, many researchers utilized uncertainty as the metric to detect anomalies.
Earlier methods used simple statistics~\cite{hendrycks2016baseline,lee2017training} like softmax probability or utilized Bayesian neural networks with MC dropout~\cite{kendall2017uncertainties,lakshminarayanan2017simple,mukhoti2018evaluating} to estimate pixel-wise uncertainty.
While those methods have low computational costs, they show an unsatisfactory accuracy in detecting anomalies~\cite{lis2019detecting}. 
The other line of work detected anomalies based on input reconstruction and prototype learning~\cite{di2021pixel,vojir2021road,xia2020synthesize,fu2023evolving,jezequel2023efficient,wu2024unsupervised}. 
Usually, an auto-encoder network is incorporated to reconstruct the original image using the predicted semantic segmentation map, and this method detects anomalies based on the visual difference between the original image and the reconstructed image.
However, the complex computation needed by the auto-encoder makes it inefficient for real-time applications. 

\textbf{Anomalous training with anomaly data.} Previous research has proven that utilizing anomalous training data, also called outlier exposure (OE), to retrain the model could greatly enhance the performance of anomaly detection~\cite{chan2021entropy,tian2022pixel,zaheer2022stabilizing}. ~\cite{hendrycks2018deep} is one of the first methods utilizing the OE from CIFAR-10 for image-level anomaly detection. Later,~\cite{di2021pixel} used the void classes in the Cityscapes as the OE, and~\cite{chan2021entropy} directly copy objects from COCO and paste them into the Cityscapes to synthesize the anomalous data.
However, those "copy-paste" based synthesizing methods never consider the harmonization between the anomaly and the image, for example, the position, size, and illumination relationship, which makes the synthesized image unrealistic. 
Meanwhile, using the objects from another close-set dataset lacks the diversity to handle the open-world challenge.
To solve that, this work thus proposes a diffusion-based anomaly data synthesizer that is capable of automatically generating a wide range of diverse and realistic anomalous images to enhance existing anomalous training.  

\textbf{Energy-based learning and mask-based classification.} The energy-based learning aims to minimize the energy of anomalies while keeping the energy of inlier classes at a high level~\cite{lecun2006tutorial}. In~\cite{tian2022pixel}, the energy is estimated by calculating the \textit{logsumexp} on the model's output~\cite{liu2020energy}, and the model is trained to minimize the energy of anomalies~\cite{chan2021entropy}.
While retraining the model makes a good improvement in detecting the anomaly, it can degrade the model's segmentation performance in inlier classes.
Later,~\cite{liu2023residual} introduced an extra residual pattern learning (RPL) module to detect the anomaly while maintaining the model's performance in segmenting inlier classes. During training, the parameters of the original model are frozen and the RPL module is trained to minimize the energy of anomalies.
In~\cite{nayal2023rba,rai2024mask2anomaly,rai2023unmasking}, a masked-based method were proposed to conduct anomaly segmentation by shifting the task from a per-pixel classification to a mask classification.
While those methods achieved good performance under existing benchmarks, they experienced a significant performance drop when faced with challenging or unexpected anomalies. 
Meanwhile, those methods are vulnerable to complex and adverse weather scenarios, as shown in \figurename~\ref{fig:visualized result}, that easily mislead the model to make a false positive prediction. 

To enhance the robustness of existing models under such intricate scenarios, this paper proposed a more effective energy-entropy learning strategy. 
By introducing a binary cross-entropy loss to enlarge the energy gap between the misleading inliers and outliers and integrating the complementary information of energy and softmax entropy to derive a more robust anomaly score, the model is trained to gain better performance and generalizability for intricate open-world challenges.
Benefiting from the development of text2image generative models~\cite{nichol2021glide,ramesh2022hierarchical,yu2022scaling,fan2022frido}, we present the first attempt to utilize the powerful text2image generative model~\cite{rombach2022high} to generate anomalous objects for open-world semantic segmentation, improving the copy-paste method and enabling the automatic generation of diverse and realistic anomalous images to enhance the anomalous training.
\begin{table*}[t]
    \centering
    \caption{Comparison of different anomaly segmentation benchmarks.}
    \footnotesize
    \resizebox{0.98\textwidth}{!}{\begin{tabular}{lcccccc}
        \toprule
        \textbf{Dataset} & \textbf{Anomaly pixels} & \textbf{Benchmark size} & \textbf{Landforms} & \textbf{Anomalies} & \textbf{Weather conditions}  \\
        \midrule
        LostAndFound test~\cite{pinggera2016lost} & 0.12\% & 1203 & 1 & 9 & clear  \\
        Fishyscapes L\&F val~\cite{blum2019fishyscapes} & 0.23\% & 373 & 1 & 7 & clear \\
        SMYIC-RoadObstacle21 test~\cite{chan2021segmentmeifyoucan} & 0.12\% & 327 & 2 & 31 & clear, frosty, nighttime  \\
        RoadAnomaly~\cite{lis2019detecting} & 9.85\% & 60 & 3 & 21 & clear  \\
        SMYIC-RoadAnomaly21 test~\cite{chan2021segmentmeifyoucan} & 13.83\% & 100 & 4 & 26 & clear  \\
        Our ComsAmy & 13.02\% & 468 & 7 & 48 & clear, riany, nighttime, frosty, foggy, blizzard, sandstorm  \\
        \bottomrule
    \end{tabular}}
    \label{tab:anomaly-dataset}
\end{table*}

\section{The ComsAmy Benchmark}
The ComsAmy benchmark aims to provide a comprehensive and reliable platform for evaluating the robustness of anomaly segmentation models within open-world environments. 
Given the inherent complexity of open-world scenarios and the diverse nature of out-of-distribution (OoD) objects encountered therein, our benchmark is carefully constructed to include a wide variety of landform types, weather conditions, and an extensive set of anomaly categories. 
This meticulous design ensures a more precise and trustworthy assessment of model performance on open-world anomaly segmentation tasks.

\begin{figure}[t]
  \centering
   \includegraphics[width=1\linewidth]{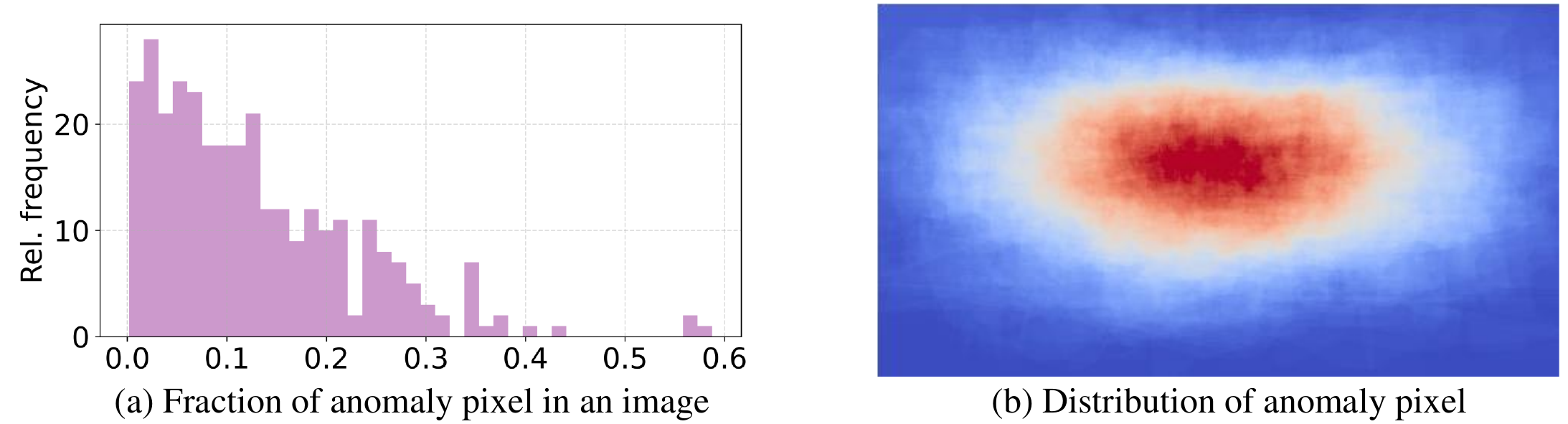}
   \caption{The fraction and distribution of annotated anomalous pixels within the image over the entire dataset.}
   \label{fig:dataset}
\end{figure}

To achieve these objectives, we initially gathered a dataset comprising 468 images sourced from the internet. 
These images capture a wide variety of anomalies under diverse weather conditions, including 250 images under clear weather, and 218 images from adverse weather such as snow, rain, fog, nighttime, blizzards, and sandstorms.
To further enhance the diversity of driving scenarios, the benchmark incorporates a wide variety of landform types, including highways, urban roads, country roads, forests, deserts, polar regions, and grasslands.
Moreover, our benchmark is specifically designed to cover an extensive range of anomalies, ensuring a comprehensive evaluation of anomaly segmentation models.
\tablename~\ref{tab:anomaly-dataset} presents an overall comparison between the proposed ComsAmy benchmark and existing benchmarks, highlighting that ComsAmy stands as the most comprehensive benchmark to cover the complex open-world scenarios with the most extraordinary diversity in landforms, anomalies, and weather conditions.
\figurename~\ref{fig:dataset} presents the proportion and spatial distribution of anomalous pixels across images within our dataset, revealing significant variability in the size and location of anomalous objects. 
This highlights the complexity and richness of the dataset, making it well-suited for evaluating anomaly detection models under diverse and challenging conditions.

\textbf{Label policy.} The ComsAmy benchmark provides pixel-level annotations for three categories: anomaly, non-anomaly, and void. 
The difference between anomaly and non-anomaly primarily follows definitions from the Cityscapes dataset, which provides fine-grained annotations for the 19 types of common objects found on roads. 
Objects in our benchmark are classified as anomalies if they fulfill three conditions: 1) the object does not belong to any of the 19 classes defined by Cityscapes; 
2) the object is either located within the driving area or could potentially move into the driving area (e.g., a billboard positioned by the roadside is not labeled as an anomaly since it does not obstruct the driving area);
and 3) the object is sufficiently large or heavy that a collision would impact the vehicle's driving (e.g., minor debris and small twigs are excluded from anomalies).
Objects meeting only the first condition but failing to satisfy either of the remaining two conditions are labeled under the void class.

\textbf{Evaluation metrics.} The pixel-level evaluation is considered in our benchmark. 
Following previous work~\cite{blum2019fishyscapes,chan2021segmentmeifyoucan,lis2019detecting}, the Area under the Precision-Recall Curve (AuPRC) and False Positive Rate at 95\% True Positive Rate (FPR95) are considered as the metrics to evaluate the model's ability for anomaly segmentation.
AuPRC is a robust metric commonly utilized to evaluate anomaly detection models, particularly in scenarios where the anomalous class constitutes a small proportion of the dataset. 
It effectively captures the trade-off between precision and recall across all classification thresholds, making it suitable for evaluating performance on imbalanced datasets, typically in anomaly segmentation tasks.
Higher AuPRC values correspond to better anomaly detection capabilities.
FPR95 quantifies the proportion of non-anomaly pixels incorrectly classified as anomalies when the detection system correctly identifies 95\% of anomalous pixels. 
Lower values of FPR95 indicate superior performance in effectively distinguishing anomalous objects from normal objects at high recall levels.
Any prediction that is within the region of void class is not counted as a false positive prediction.



\begin{figure*}[t]
  \centering
   \includegraphics[width=1\linewidth]{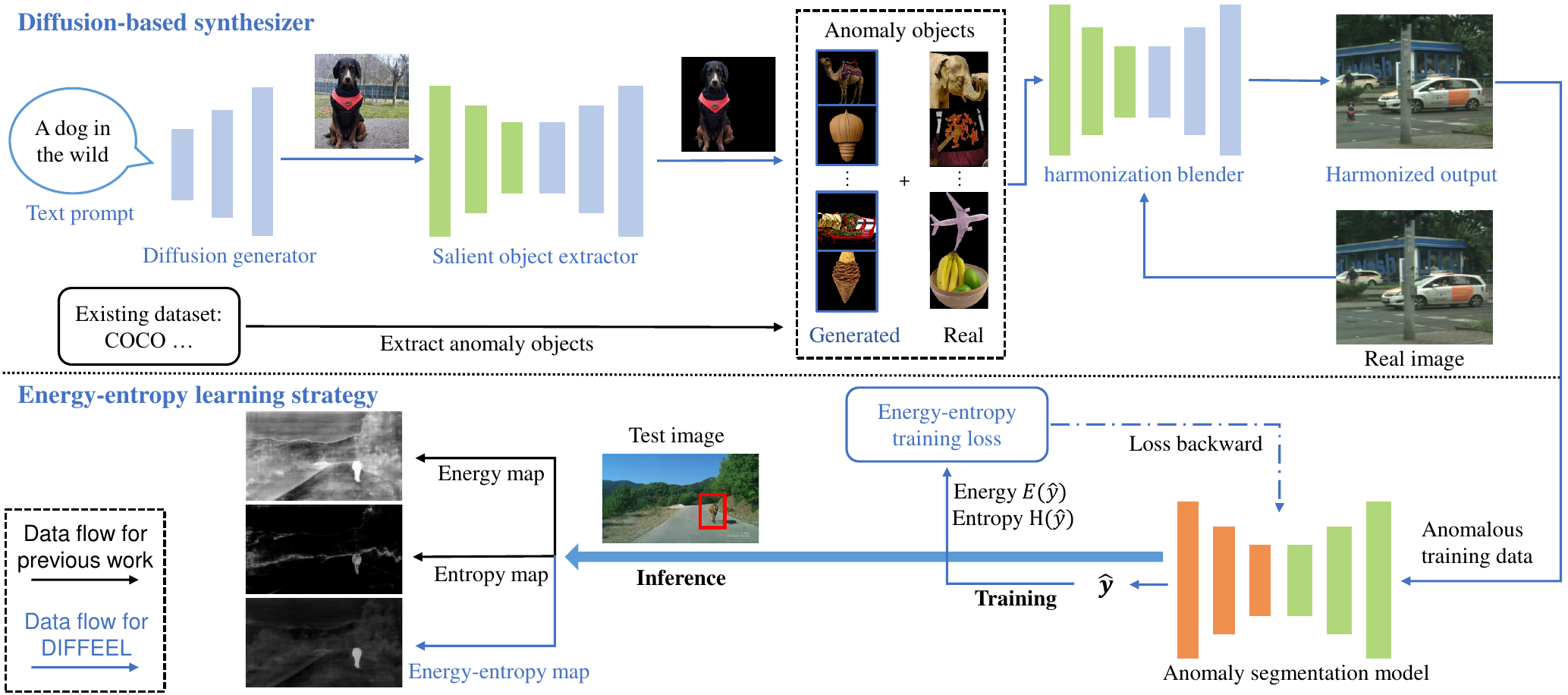}
   \caption{The structure of the proposed diffusion-based anomaly data synthesizer and energy-entropy learning strategy. The object generator can create a large diversity of anomalous objects, and the salient object extractor will extract the precise mask of the anomalous object to enhance the variety of the anomalies. The harmonization blender will blend the anomalous object with a harmonious size, position, and illumination. 
   The EEL strategy influences the training and inference stages, which can improve model generalization and performance in intricate open-world scenarios.  }
   \label{fig:flow_chat_df}
\end{figure*}
\section{Methodology}
This section includes two parts. 
The first part briefly introduces the anomalous training dataset and presents the details of our diffusion-based anomaly training data synthesizer.
The second part presents the details of the proposed energy-entropy learning (EEL) strategy.
\subsection{Diffusion-based Anomaly Data Synthesizer}
Assume that we have an inlier training dataset (\eg, Cityscapes~\cite{cordts2016cityscapes}) denoted by ${D^{in}} = \{ {x^{in}},y^{in}\}$, where $x^{in} \in X \subset {\mathbb{R}^{H \times W \times 3}}$ represents the $H \times W$ training image with $3$ color channels, and $y^{in} \in {Y^{in}} \subset {\mathbb{N}^{H \times W }}$ represents the $H \times W$ pixel-wise label for $C$ classes. 
Let us assume another dataset with out-of-distribution (OoD) objects denoted by ${D^{out}} =  \{ {x^{out}},y^{out}\}$, where $x^{out} \in X^{out} \subset {\mathbb{R}^{H' \times W' \times 3}}$ represents the image of the OoD object, and ${y^{out}} \in {Y}^{out} \subset {\{ 0,1\} ^{H' \times W'}}$ represents the pixel-wise binary mask of the OoD object with $1$ denoting the object pixel and $0$ denotes the background.
Following previous work~\cite{grcic2022densehybrid,tian2022pixel}, the anomaly training dataset ${D^{amy}} = \{ {x^{amy}},y^{amy}\}$ that contains the OoD object is generated by copying the anomaly objects from ${D^{out}}$ and pasting them into ${D^{in}}$.
This process is formulated as :
\begin{equation}
\begin{array}{l}
\setlength{\jot}{20pt}
{x^{amy}} = Pad (1 - {y}^{out}) \odot {x^{in}} + Pad ({y}^{out} \odot {x}^{out}),
\\
{y^{amy}} = Pad (1 - {y}^{out}) \odot {y^{in}} + Pad({y}^{out}) * P,
\end{array}
\label{eq:anomaly}
\end{equation} 
where $Pad$ represents the zero-padding process to the spatial resolution of $H\times W$. 
The element-wise multiplication $\odot$ with $Pad (1-{y}^{out})$ will make the pixels in the original image that overlap with the anomaly as 0 and others remain the same. 
$P$ is an integer larger than $C$ to distinguish the anomaly from inlier classes (\eg $P=20$ if there are 19 classes in the Cityscapes dataset). 
Equation~\ref{eq:anomaly} will blend the OoD object with the original image and update the label according to the OoD mask.
%
\begin{figure}[t]
  \centering
   \includegraphics[width=1\linewidth]{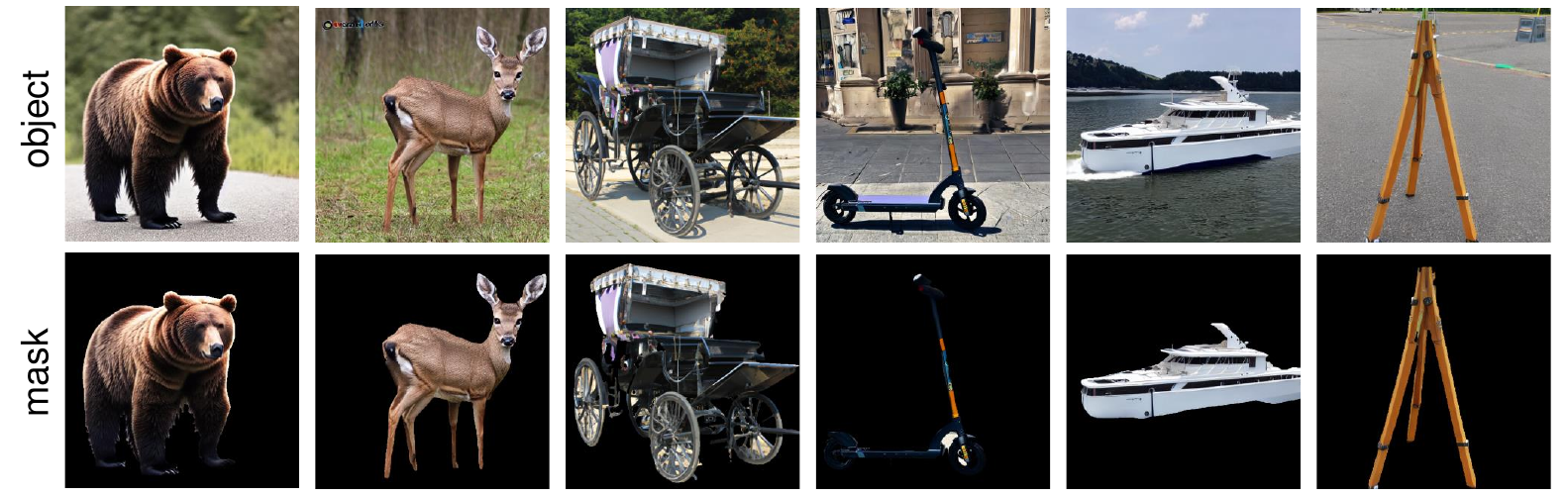}
   \caption{The diffusion generated anomalous objects and the extracted masks.}
   \label{fig:diff_objs}
\end{figure}

Unlike previous work ~\cite{grcic2022densehybrid,tian2022pixel} that take the OoD objects from another closed-set dataset such as COCO~\cite{lin2014microsoft}, enhanced by the powerful text2image generator, the $x^{out}$ in our dataset synthesizer is more diverse. 
Meanwhile, $x^{out}$ will be processed for the harmonization to produce a more realistic synthesized image. 
The structure of the diffusion-based anomaly data synthesizer is depicted in \figurename~\ref{fig:flow_chat_df}. 
For generating anomalous objects, we employ the stable diffusion text-to-image model~\cite{rombach2022high}, which allows for the creation of diverse anomalous objects guided by textual prompts.
A representative example of text description is: "One dog (type of the object) in the wild (the background)" to control the type of the generated object and its background.
Then, the advanced salient object extractor~\cite{qin2020u2} is utilized to extract the generated object from its background to facilitate the blending and harmonization process.
Some examples of generated objects and extracted masks are shown in the \figurename~\ref{fig:diff_objs}.
Subsequently, the anomalous object, its corresponding mask, and the real images from Cityscapes are concatenated and input into the harmonization blender~\cite{cong2020dovenet} for seamless integration.
The harmonization blender initially searches the entire image for an optimal placement position. Upon identifying this position, it performs necessary post-processing steps, including adjustments to the object's size and illumination, and subsequently employs Equation~\ref{eq:anomaly} to generate the anomalous image.
\begin{figure}[t]
  \centering
   \includegraphics[width=1\linewidth]{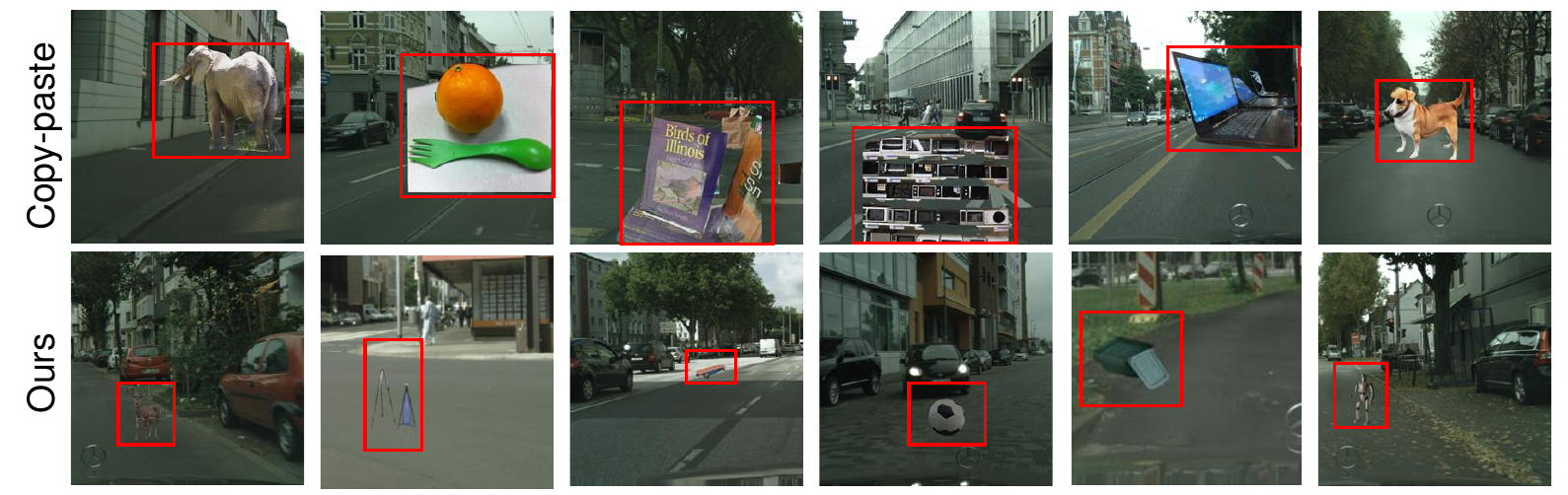}
   \caption{The comparison of anomaly images generated by the "copy-paste" method and ours. The anomaly is marked by the red bounding box.}
   \label{fig:cp-harm}
\end{figure}

As shown in the upper right of \figurename~\ref{fig:flow_chat_df} and \figurename~\ref{fig:cp-harm}, anomalous images created using the previous copy-paste method often exhibit inconsistencies between the inserted object and the background, particularly regarding illumination, positioning, and scaling. These discrepancies result in an unnatural appearance, characterized by a visibly inharmonious relationship between the object and its surrounding context.
Consequently, anomaly segmentation models trained on such artificially generated data may erroneously focus on these unnatural relationships, impairing their ability to accurately detect anomalies in real-world scenarios.
In contrast, the anomalous images produced by our proposed harmonization blender demonstrate a significantly more realistic integration of objects into the background, thereby facilitating improved anomaly detection performance in practical applications.


\subsection{Energy-entropy Learning Strategy}

\begin{figure}[t]
  \centering
   \includegraphics[width=0.95\linewidth]{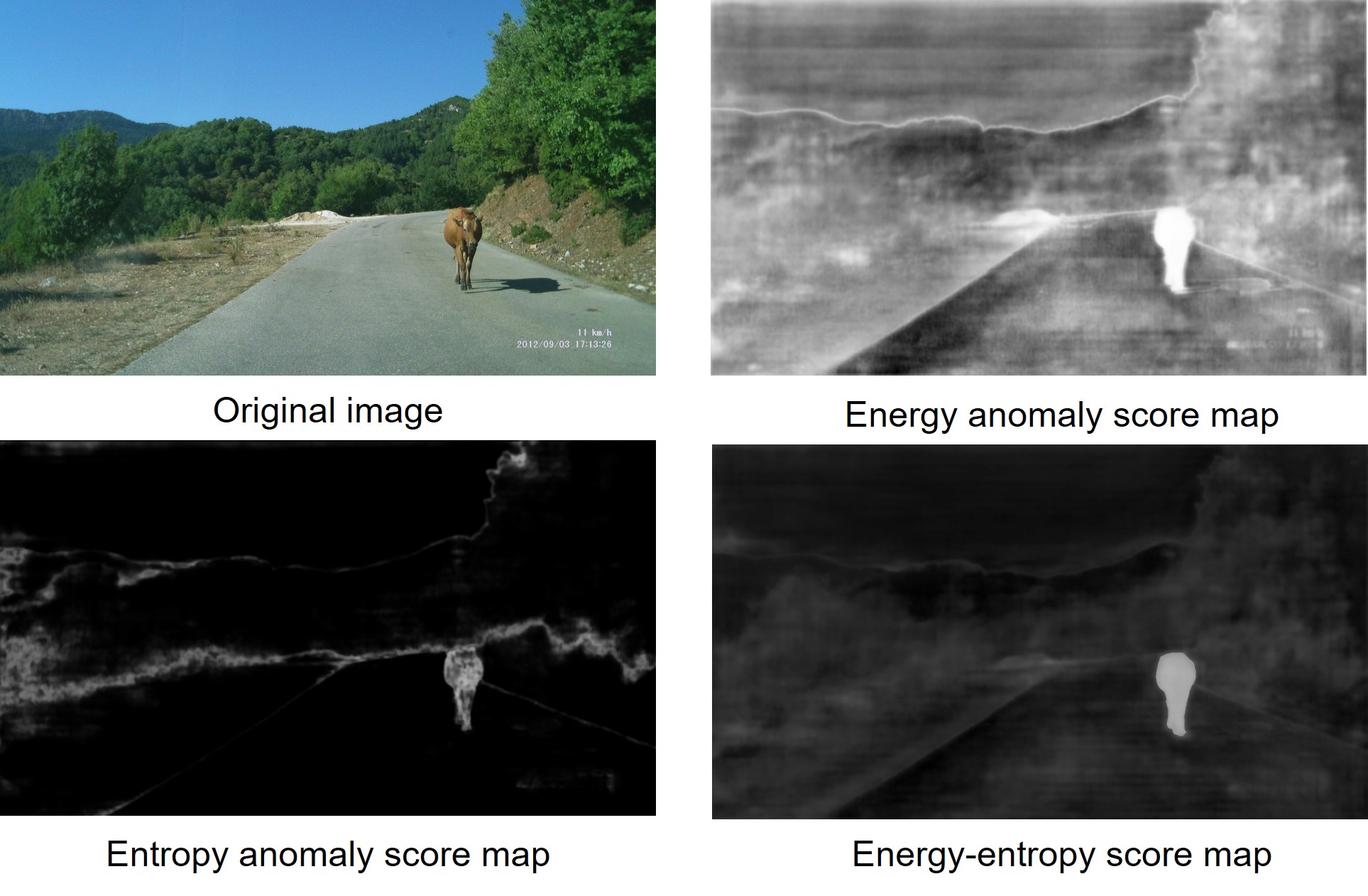}
   \caption{The comparison between energy and entropy anomaly score map. The grey level represents the anomaly scores. 
   The combination of energy and entropy provides a more robust anomaly score for detecting anomalies.}
   \label{fig:energy-entropy}
\end{figure}
\textbf{Anomaly segmentation models.} 
Let $x \subset {\mathbb{R}^{H \times W \times 3}}$ be the input image and $y \subset {\mathbb{N}^{H \times W }}$ be the corresponding semantic label with $C$ classes.
Existing anomaly segmentation models can be broadly categorized into pixel-wise and mask-wise segmentation. 
The pixel-wise anomaly segmentation model~\cite{liu2023residual,tian2022pixel,vojivr2024pixood,tu2024self,sodano2024open,shoeb2024have} generally outputs the prediction logits $\hat{y} \subset {\mathbb{R}^{C \times H \times W}}$, and an intuitive approach to detecting anomalies involves applying the maximum softmax probability (MSP)~\cite{hendrycks2016baseline} on top of the prediction logists, which is:
\begin{equation}
    \mathbb{S}_{amy}(x)=1- \max (Softmax(\hat{y} )).
    \label{eq: pixel-wise anomaly score}
\end{equation}
The mask-wise anomaly segmentation model~\cite{rai2023unmasking,rai2024mask2anomaly,nayal2023rba,delic2024outlier} generally outputs $N$ class masks $m \subset \mathbb{R}^{N \times H\times W}$ and its corresponding class scores $c \subset \mathbb{R}^{ N\times C}$.
The anomaly score in mask-wise methods is computed as follows:
\begin{equation}
    \mathbb{S}_{amy}(x)=1- \max (Softmax(c)^{T} \cdot Sgm(m) ),
    \label{eq: mask-wise anomaly score}
\end{equation}
where $Sgm$ is the Sigmoid function.

\textbf{Energy-entropy anomaly score.} In~\cite{tian2022pixel,liu2023residual}, the energy is utilized to derive the anomaly score for pixel-wise anomaly segmentation models, and pixels with low energy are assigned high anomaly scores. 
The energy $\mathrm {E}$ is calculated by the $logsumexp$ of the model's prediction, shown as follows:
\begin{equation}
\mathrm {E}({\hat y}) = \log \left( {\sum\limits_{i = 1}^C {\exp ({{\hat y}_i})} } \right),
\label{eq:energy}
\end{equation} 
where $\exp$ is the exponential function and $C$ represents all inlier classes.
For mask-wise anomaly segmentation models, we can calculate the prediction logits by: $\hat{y}= c^T \cdot Sigmoid(m)$.
The core idea of using energy as the metric is that anomalies contain some unknown features that cannot activate the trained neural network. 
However, unfamiliar or challenging inlier objects may also include some inactive features, leading to an overall low feature response. 
Additionally, the energy function disregards the relative relationship of outputs between different classes, which can result in a pixel with a high softmax confidence being classified as an anomaly. 
The softmax entropy $\mathrm {H}$, defined in Equation~\ref{eq:entropy} with ${\hat p_i} = softmax {(\hat y)_i}$, evaluates the prediction certainty across all classes and shows good performance in recognizing inlier objects.
\begin{equation}
\mathrm{H}({\hat y}) = \sum\limits_{i = 1}^C {{-\hat p_i}\log ({\hat p_i})}.
\label{eq:entropy}
\end{equation} 
As shown in \figurename~\ref{fig:energy-entropy}, many inlier background pixels are assigned high energy anomaly scores while getting low entropy anomaly scores. 
So, in this paper, we utilize the complementary information from both energy and entropy to generate a more robust anomaly score ${\mathbb{S}^{eel}_{amy}}$, defined in Equation~\ref{eq:score}.  
\begin{equation}
{\mathbb{S}^{eel}_{amy}} =\alpha  \cdot \mathrm{H}({\hat y}) -\mathrm{E}({\hat y}) .
\label{eq:score}
\end{equation} 

\begin{figure}[t]
  \centering
   \includegraphics[width=0.85\linewidth]{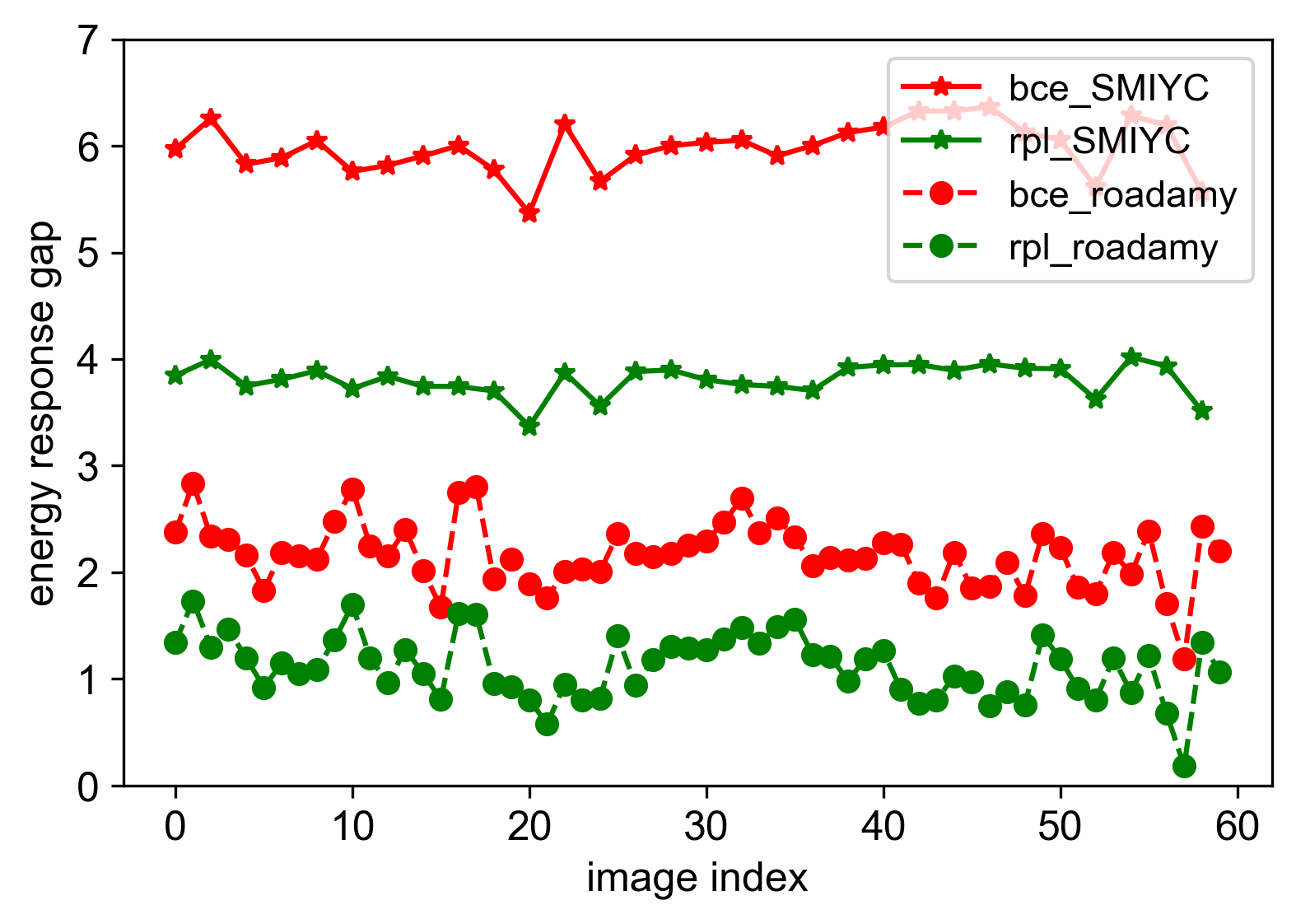}
   \caption{The energy response gap between high energy response anomalies and low energy response inliers of models trained on our binary energy-entropy learning loss and rpl loss in SMIYC obstacle and RoadAnomaly validation sets.}
   \label{fig:energy_gap}
\end{figure}

\textbf{Energy-entropy training loss.} In~\cite{liu2023residual}, the energy response of anomalies is optimized based on a linear function.
Instead, we utilized a sigmoid function to turn the energy response into the anomaly probability and optimize this probability with a binary cross-entropy loss, which is defined in Equation~\ref{eq:EEL_loss}:
\begin{equation}
\begin{small}
\begin{aligned}
\!\!\!\!\mathcal{L} ^{eel} \!\!=\!\! \sum\limits_{(x,y) \in {D^{amy}}} \!\!\!\! {\mathbb{E}_{\scriptstyle\omega  \in \Omega \hfill\atop
\scriptstyle{m_\omega } = 1\hfill}}\!\!\!\left( ( {-\log \left( {Sgm\left( {-\mathrm{E}({{\hat y}_\omega })} \right)} \right)} -\alpha  \cdot \mathrm{H}(\hat y_\omega)\right) &\\- {\mathbb{E}_{\scriptstyle\omega  \in \Omega \hfill\atop
\scriptstyle{m_\omega } = 0\hfill}}\left( {\log \left( {1 - Sgm\left( {-\mathrm{E}({{\hat y}_\omega })} \right)} \right)}+\alpha  \cdot \mathrm{H}(\hat y_\omega)) \right),&
\label{eq:EEL_loss}
\end{aligned}
\end{small}
\end{equation} 
where $\Omega$ is the image lattice of size $H\times W$. 
$m_\omega \subset {\{ 0,1\} ^{H \times W}}$ is the mask for anomaly segmentation, where 1 denotes anomalies and 0 denotes inlier classes.
$\mathbf{1}\{\cdot\}$ returns 1 when the interior condition is met, otherwise 0. 
$Smg$ is the sigmoid function.
Compared with the energy-based optimization in~\cite{liu2023residual}, the advantages of Equation~\ref{eq:EEL_loss} are that firstly, $\mathcal{L} ^{eel}$ considers the energy response for both the inlier and outlier. 
Moreover, $\mathcal{L} ^{eel}$ demonstrates a non-linear gradient towards the energy response: greater penalties will be imposed on misleading anomalies and inliers that exhibit less favorable energy responses. 
\figurename~\ref{fig:energy_gap} gives a comparative analysis of the average energy gap between the highest energy anomalies of the top 25\% and the lowest 25\% energy inliers, indicating that our $\mathcal{L} ^{eel}$ can better discriminate challenging inliers and anomalies than the linear optimization in~\cite{liu2023residual}.

\textbf{Combination of EEL strategy with existing methods.} By integrating the proposed EEL loss into the training process, the EEL learning strategy can serve as a "plug \& play" method to enhance both pixel-wise and mask-wise anomaly segmentation models. 
To verify the effectiveness of the proposed EEL learning strategy, we combine EEL with two SOTA anomaly segmentation models: the pixel-wise RPL model~\cite{liu2023residual}, and the mask-wise Mask2Anomaly~\cite{rai2024mask2anomaly} model.  
Specifically, for the RPL model~\cite{liu2023residual}, we replace its original energy-based anomaly loss with the proposed EEL loss $\mathcal{L} ^{eel}$ while retaining its consistency loss $\mathcal{L} ^{cons}$, which is defined as follows:  
\begin{equation}
{\mathcal{L} } = \mathcal{L} ^{cons} + \lambda  \cdot\mathcal{L} ^{eel},
\label{eq:loss}
\end{equation} 
where $\mathcal{L} ^{eel}$ aims to make the model detect the anomalies and $\mathcal{L} ^{cons}$ aims to make the model's prediction ${\hat y}$ consistent with the original prediction ${\tilde y}$ among the inlier classes. 
The $\mathcal{L} ^{cons}$ is defined as:
\begin{equation}
\begin{aligned}
\mathcal{L} ^{cons} = \sum\limits_{(x,y) \in {D^{amy}}} \sum\limits_{\omega  \in \Omega } \mathbf{1}\{ {m_\omega } = 0\} ({\mathrm{H}_{CE}}({{\tilde y}_\omega },{{\hat y}_\omega })&\\ + {D_{KL}}({{\tilde y}_\omega }\parallel {{\hat y}_\omega }) + {D_{MSE}}(\mathrm{H}({{\tilde y}_\omega }),\mathrm{H}({{\hat y}_\omega }))),&
\label{eq:lcons}
\end{aligned}
\end{equation}  
where ${H_{CE}}$ and ${D_{KL}}$ respectively denote the cross-entropy and Kullback-Leibler divergence. $D_{MSE}$ calculates the mean square error between the softmax entropy $H$ of $\tilde y$ and $\hat y$. 
Similarly, for the Mask2Anomaly model~\cite{rai2024mask2anomaly}, we enhance the training by substituting its original mask contrastive learning loss with our proposed EEL loss while retaining other constraint terms. The revised loss function is expressed as:
\begin{equation}
{\mathcal{L} } = \mathcal{L} ^{masks} + \lambda^{ce}\cdot\mathcal{L} ^{ce} + \lambda\cdot\mathcal{L} ^{eel},
\label{eq:m2a_loss}
\end{equation} 
where the $\mathcal{L} ^{ce}$ and $\mathcal{L} ^{masks}$ are constraint terms intended to maintain segmentation accuracy for inlier classes. 
Further details of $\mathcal{L} ^{ce}$ and $\mathcal{L} ^{masks}$ are provided in Section IV of~\cite{rai2024mask2anomaly}.

\begin{table*}[t]
\renewcommand\arraystretch{1.1}
  \centering
  \caption{The pixel-wise evaluation results of SOTA anomalous segmentation models on our ComsAmy benchmark.}\label{tab:experiment ComsAmy}

  \scalebox{1}{
  \setlength{\tabcolsep}{3mm}{\begin{tabular}{c|cc|cc|cc}
    \hline
    \multirow{2}*{Method} &  \multicolumn{2}{c|}{Clear weather} &  \multicolumn{2}{c|}{Adverse weather} &  \multicolumn{2}{c}{Whole dataset} \\
    \cline{2-7}
    & \thead{$\rm{AUPRC}~\uparrow$} & \thead{$\rm{FPR_{95}}\downarrow$} & \thead{$\rm{AUPRC}~\uparrow$} & \thead{$\rm{FPR_{95}}\downarrow$}
    & \thead{$\rm{AUPRC}~\uparrow$} & \thead{$\rm{FPR_{95}}\downarrow$}\\
    \hline
    Maximum Softmax~\!{\tiny \textcolor{gray}{[ICLR17]}}  &32.85 &68.32 &19.28 &83.06 &25.79 &75.31  \\    
    DenseHybrid~\!{\tiny \textcolor{gray}{[ECCV22]}} &29.42 &96.27 &24.96 & 94.84 &28.02 & 95.64 \\ 
    Meta\_OoD~\!{\tiny \textcolor{gray}{[ICCV21]}} &55.16 &58.51 &49.82 & 55.00 & 56.52 & 54.90 \\ 
    PEBAL~\!{\tiny \textcolor{gray}{[ECCV22]}} &40.38 &55.41 &21.33 & 84.16 & 31.50 & 68.70 \\ 
    RPL~\!{\tiny \textcolor{gray}{[ICCV23]}} &65.42 &43.60 &49.67& 61.42 & 59.43  & 50.73 \\ 
    M2A~\!{\tiny \textcolor{gray}{[TPAMI24]}} &68.87 &62.03 &47.30 &59.67 &60.67 & 60.21 \\ 
    \hline
    RPL+DiffEEL~\!{\tiny \textcolor{gray}{[Ours]}} &72.49 &32.86 & \textbf{54.70} & \textbf{47.32} & \textbf{66.12} & \textbf{38.76}\\
    M2A+DiffEEL~\!{\tiny \textcolor{gray}{[Ours]}} &\textbf{72.85} &\textbf{31.52} &52.34 &54.09 & 64.75 & 42.53 \\
    \hline
  \end{tabular}}}
\end{table*}
\section{Experiments}
\subsection{Experiment Setup}
\textbf{Evaluation benchmarks.} Consistent with the prior work~\cite{liu2023residual,tian2022pixel,rai2024mask2anomaly,rai2023unmasking,nayal2023rba}, we evaluate our proposed method on three public benchmarks, namely Fishyscapes, SMIYC, and RoadAnomaly, and our proposed ComsAmy benchmark. 
The Fishyscapes~\cite{blum2021fishyscapes,blum2019fishyscapes} contains two subsets called Fishyscapes Lost $\&$ Found (FS-L$\&$F), and Fishyscapes Static (FS-Static).
The FS-L$\&$F validation set contains 100 images with anomalous obstacles from the LostAndFound dataset~\cite{pinggera2016lost}.
The FS-Static validation set consists of 30 urban scene images with artificially synthesized anomalies.
The RoadAnomaly~\cite{lis2019detecting} benchmark comprises 60 images of potential hazards on the road, such as animals, rocks, and other obstacles. 
The SMIYC~\cite{chan2021segmentmeifyoucan} benchmark contains two tracks: the anomaly and the obstacle tracks.
The Anomaly track offers a validation set of 10 images and an online test set of 100 images, including animals and unidentified vehicles. The Obstacle track provides a validation set of 30 images and an online test set of 327 images.
This track focuses on obstacles on the road, such as stuffed toys and tree stumps. 
The ComsAmy benchmark stands as the most challenging benchmark that comprises 468 anomalous images that encompass 48 distinct anomaly types distributed across 7 unique landforms and captured under 7 varying weather conditions. 
This benchmark aims to offer a robust and comprehensive assessment of anomaly segmentation models under complex open-set scenarios.

\textbf{Training set.} The original training images come from the Cityscapes dataset~\cite{cordts2016cityscapes} containing 2,975 images for training, 500 images for validation, and 1,500 images for testing. The anomalous objects come from our diffusion-based object generator and COCO dataset~\cite{lin2014microsoft}. Compared with the previous work~\cite{liu2023residual} that obtains 46,751 anomalous objects from COCO, we use a comparable amount of 44,500 anomalous objects with 25,051 objects generated by the diffusion-based generator and 19,449 objects obtained from COCO. Improvements are that our anomalies are more diverse due to the utilization of the diffusion generator and our synthesized images are more realistic by using the harmonization blender. 

\textbf{Model implementation details.} To combine our proposed DiffEEL with the RPL method, we follow~\cite{liu2023residual,tian2022pixel} that utilizes the architecture provided by DeepLabV3+ with WiderResNet38~\cite{zhu2019improving} as the backbone and takes the pre-trained model on Cityscapes provided by~\cite{zhu2019improving}. 
When re-training on the anomaly data, we freeze the parameters of the original model and update the parameters of the RPL module, which can guarantee the model's semantic segmentation performance in the inlier classes. The initial learning rate is $5{e^{ - 5}}$ with poly learning rate decay, and the total training epochs are 30.
For the Mask2Anomaly model, we follow~\cite {rai2023unmasking,rai2024mask2anomaly} that utilizes the ResNet-50~\cite{he2016deep} encoder, and its weights are initialized from a model that is pre-trained with barlow-twins~\cite{zbontar2021barlow} self-supervision on ImageNet~\cite{deng2009imagenet}. 
The multi-scale deformable attention Transformer(MSDeformAttn)~\cite{zhudeformable} is utilized as the pixel decoder. it gives feature maps at 1/8, 1/16, and 1/32 resolution, providing image features to the transformer decoder layers. 
The transformer decoder is adopted from Cheng et al.~\cite{cheng2022masked} and consists of 9 layers with 100 queries.
We train the network for 4,000 iteration using AdamW with a weight decay of 0.05, a learning rate of 1e-5, and a batch size of 8.
For hyperparameters, we set $\alpha$ in Equation~\ref{eq:score} as 1, $\lambda$ in Equation~\ref{eq:loss} as 0.05 and $\lambda^{ce}$ in Equation~\ref{eq:m2a_loss} as 1.

\textbf{Evaluation metrics:} Following previous works~\cite{blum2021fishyscapes, chan2021entropy}, we report the area under the receiver operating \ the area under the precision-recall curve ($\rm{AUPRC}$), and the false positive rate at a true positive rate of 95\% ($\rm{FPR}_{95}$).

\begin{table*}[t]
\renewcommand\arraystretch{1.1}
  \centering
  \caption{{The comparison results on the validation set of FS-L\&F,  FS-Static, SMIYC-obstacle, and RoadAnomaly Benchmarks. 
  }}\label{tab:experiment FS_SMIYC}
  \resizebox{\textwidth}{!}{
  \begin{tabular}{c|cc|cc|cc|cc|cc}
    \hline
    \multirow{2}*{Method} & \multicolumn{2}{c|}{FS-L\&F}&\multicolumn{2}{c|}{FS-Static} & \multicolumn{2}{c|}{SMIYC-obstacle}&\multicolumn{2}{c|}{RoadAnomaly} &\multicolumn{2}{c}{Average}\\
    \cline{2-11}
    & \thead{$\rm{AUPRC}~\uparrow$} &\thead{$\rm{FPR_{95}}\downarrow$}& \thead{$\rm{AUPRC}~\uparrow$} &\thead{$\rm{FPR_{95}}\downarrow$}& \thead{$\rm{AUPRC}~\uparrow$} &\thead{$\rm{FPR_{95}}\downarrow$}& \thead{$\rm{AUPRC}~\uparrow$} &\thead{$\rm{FPR_{95}}\downarrow$} &\thead{$\rm{\tilde{AUPRC}}~\uparrow$} &\thead{$\rm{\tilde{FPR_{95}}}\downarrow$}
    \\  
    \hline
    Maximum Softmax~\!{\tiny \textcolor{gray}{[ICLR17]}} & 40.34&10.36& 26.77 & 23.31 &43.4 &3.8&60.2 &40.4  & 42.67 & 19.46  \\    
    Synboost~\cite{di2021pixel}
    ~\!{\tiny \textcolor{gray}{[CVPR21]}} &60.58&31.02 & 66.44 &25.59 &81.4 &2.8  &41.83 & 59.72  &62.56&29.78\\ 
    SML~\cite{jung2021standardized}
    ~\!{\tiny \textcolor{gray}{[ICCV21]}} &22.74&33.49 & 66.72 & 12.14  &18.60 &91.31  &25.82& 49.74  &33.47&46.67\\ 
    Meta$\_$OoD~\cite{chan2021entropy}
    ~\!{\tiny \textcolor{gray}{[ICCV21]}}&41.31&37.69 & 72.91 &13.57  &94.14 &0.41  &- &-  &- &- \\ 
    DenseHybrid~\cite{grcic2022densehybrid}
    ~\!{\tiny \textcolor{gray}{[ECCV22]}} &69.79&5.09  & 76.23 &4.17 &89.49 &0.71  &- & - &-&-\\ 
    PEBAL~\cite{tian2022pixel}
    ~\!{\tiny \textcolor{gray}{[ECCV22]}} &58.81&4.76 &92.08 &1.52 &10.45 &7.92  &62.37 & 28.29  &55.93&10.62\\ 
    RPL~\cite{liu2023residual} 
    ~\!{\tiny \textcolor{gray}{[ICCV23]}}&66.73&3.66  &88.96 & 0.98 &95.20 &0.13 &66.66 & 25.73  &79.38&7.62\\ 
    RPL+CoroCL~\cite{liu2023residual} 
    ~\!{\tiny \textcolor{gray}{[ICCV23]}}&70.61 &2.52 & 92.46& 1.52 &96.91 &\textbf{0.09}   &71.61 & 17.74  &82.89& 5.46\\ 
     M2A~\cite{rai2024mask2anomaly}~\!{\tiny \textcolor{gray}{[TPAMI24]}} & 69.40& 9.41 & 90.12 &1.33 & 92.83 &0.19  &79.70 &13.45 &83.01 & 6.10
    \\
    \hline
    RPL+DiffEEL~\!{\tiny\textcolor{gray}{[Ours]}}&\textbf{77.87}& \textbf{1.12} &\textbf{94.93}&\textbf{0.55} & \textbf{97.30} &0.10  &74.22 &16.74 &\textbf{86.08} & \textbf{4.62}  \\
    M2A+DiffEEL~\!{\tiny\textcolor{gray}{[Ours]}}&77.38 & 7.62 & 88.62 & 1.17& 96.09 &0.11 &\textbf{81.54} &\textbf{12.06} &85.90 & 5.24  \\
    \hline
  \end{tabular}}
\end{table*}

\begin{figure}[t]
  \centering

   \includegraphics[width=1\linewidth]{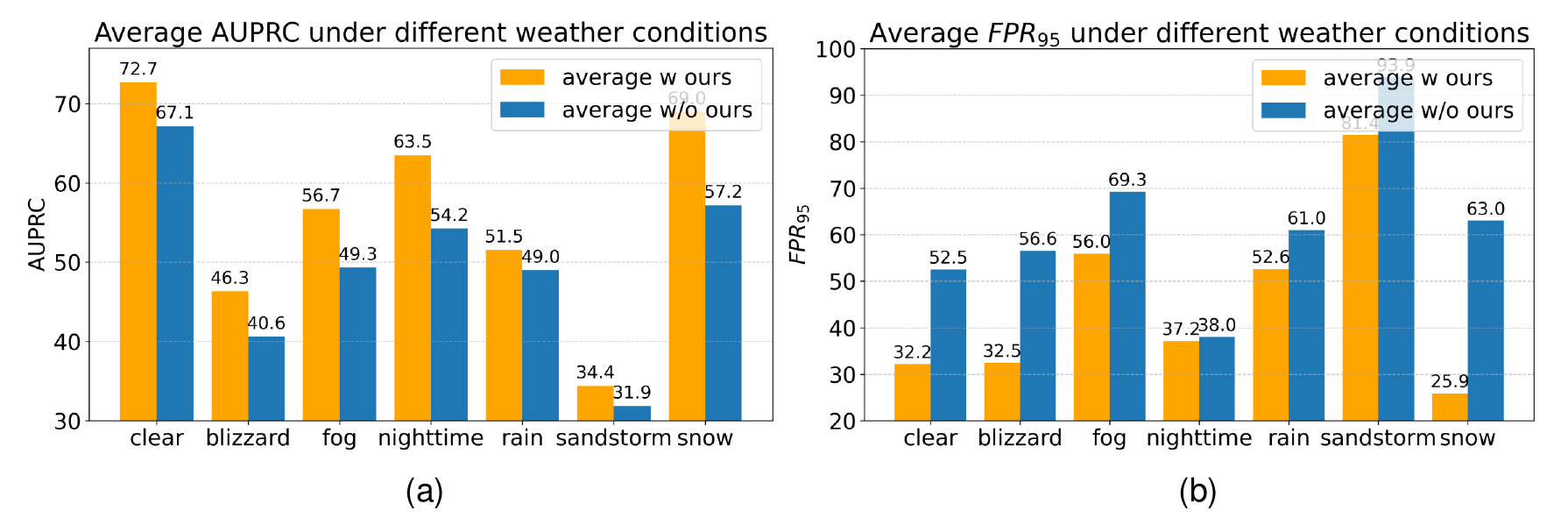}
   \caption{The performance of RPL and M2A model under different weather conditions. The bar "average w ours" is the average performance of PRL and M2A after combining with our DiffEEL, and "average w/o ours" is the average performance of their original models.}
   \label{fig:average_fpr_auprc_weather}
\end{figure}
\subsection{Experimental results on ComsAmy benchmark}
The evaluation results on our proposed ComsAmy benchmark are presented in \tablename~\ref{tab:experiment ComsAmy}.
To rigorously compare the performance of different models under clear and adverse weather conditions, we manually partitioned the benchmark into two subsets: one with 250 anomaly images captured under clear weather, and the other with 218 images captured exclusively under various adverse weather.
As shown in \tablename~\ref{tab:experiment ComsAmy}, compared to the performance under clear conditions, most evaluated methods exhibit significant degradation in $\rm{AUPRC}$ scores and a notably higher $\rm{FPR_{95}}$ when confronted with adverse weather.
These findings indicate the substantial impact of weather conditions on models' performance, highlighting the vulnerability of current anomaly segmentation models when operating in adverse environments.
To validate the effectiveness of the proposed DiffEEL strategy, we evaluate the performance of SOTA models RPL and M2A when combined with our proposed DiffEEL strategy. 
The results indicate that the proposed DiffEEL strategy can be seamlessly integrated with existing methods to bring substantial benefits on their anomaly segmentation ability.
In general, integrating DiffEEL brings an overall improvement in $\rm{AUPRC}$ of approximately 5.39\% and reduces $\rm{FPR_{95}}$ to approximately 14.83\%.

\textbf{Further discussion on performance under adverse weather:}
While integrating DiffEEL leads to significant improvements in anomaly segmentation performance, as shown in \tablename~\ref{tab:experiment ComsAmy}, a noticeable performance gap remains between the model's performance under clear and adverse weather conditions.
To better understand the impact of each individual adverse weather, we evaluate the performance of two SOTA models—RPL and M2A—under each specific weather condition.
\figurename~\ref{fig:average_fpr_auprc_weather} shows the average $\rm{AUPRC}$ and $\rm{FPR_{95}}$ for both models, before and after integrating the proposed DiffEEL. 
The results reveal that existing models exhibit a degree of robustness in conditions such as nighttime, frosty, and foggy weather. 
However, they are particularly vulnerable to sandstorms, blizzards, and rain, which lead to significant $\rm{AUPRC}$ degradation and a marked increase in $\rm{FPR_{95}}$. 
One possible explanation for this vulnerability is that adverse weather conditions, such as blizzards, severely disrupt the captured visual information, leading to great perturbations. 
Models lacking robust training to accommodate these natural perturbations are prone to making erroneous predictions. 
These findings underscore significant concerns about the performance of current anomaly segmentation models in adverse weather conditions, which are likely to be encountered in open-set environments.
\begin{table*}[ht]
  \centering
  \caption{{The comparison results on the test set of Fishscapes and SMIYC. The best result of each column is bold.}}
  \resizebox{1.0\textwidth}{!}{
  \begin{tabular}{c|cc|cc|cc|cc|cc}
    \hline
    \multirow{2}*{Method} & \multicolumn{2}{c|}{FS-L\&F}& \multicolumn{2}{c|}{FS-Static}&\multicolumn{2}{c|}{SMYIC-Anomaly}&\multicolumn{2}{c|}{SMYIC-Obstacle}&\multicolumn{2}{c}{Overall} \\
    \cline{2-11}
    & \thead{$\rm{AUPRC}~\uparrow$} &\thead{$\rm{FPR}_{95}\downarrow$}& \thead{$\rm{AUPRC}~\uparrow$} &\thead{$\rm{FPR}_{95}\downarrow$} & \thead{$\rm{AUPRC}~\uparrow$} &\thead{$\rm{FPR}_{95}\downarrow$} & \thead{$\rm{AUPRC}~\uparrow$} &\thead{$\rm{FPR}_{95}\downarrow$} 
    &\thead{$\tilde{{\rm{AUPRC}}}~\uparrow$}&\thead{${\tilde{\rm{FPR}_{95}}}~\downarrow$}\\   
    \hline
    Maximum Softmax~\cite{hendrycks2016baseline}
    {\tiny \textcolor{gray}{[ICLR17]}} & 2.93&44.83 &15.42 &39.75 &27.97 &72.05&15.72 &16.60 &3.8 & 66.98 \\    
    Synboost~\cite{di2021pixel}
    {\tiny \textcolor{gray}{[CVPR21]}} &43.22&15.79 &72.59 &18.75 &56.44 &61.86 &71.34 & 3.15 &15.51 &43.30\\ 
    Meta$\_$OoD~\cite{chan2021entropy}
    {\tiny \textcolor{gray}{[ICCV21]}}&29.96&35.14 &86.55 &8.55 &{85.47} &15.00 &85.07 &0.75 &71.76 &14.86 \\ 
    DenseHybrid~\cite{grcic2022densehybrid}
    {\tiny \textcolor{gray}{[ECCV22]}} &43.90&6.18 &72.27 &5.51 &42.05 &62.25 &80.79 & 6.02 &59.80 &20.00\\ 
    PEBAL~\cite{tian2022pixel}
    {\tiny \textcolor{gray}{[ECCV22]}} &44.17&6.18 &92.38 &1.73 &49.14 &40.82 &4.98 & 12.86 &47.66 &15.39\\ 
    M2A~\cite{rai2024mask2anomaly}{\tiny \textcolor{gray}{[TPAMI24]}}
    &46.04 &4.36 &95.20 &0.82 &\textbf{88.70} &14.60 &\textbf{93.30} &\textbf{0.20} &80.81 & 5.00\\
    RPL+CoroCL~\cite{liu2023residual} 
    {\tiny \textcolor{gray}{[ICCV23]}}&55.39 &2.27 &95.96 &0.52 & 83.49 &\textbf{11.68}  &{85.93} & {0.58} &80.19 &\textbf{3.76}\\ 
    \hline
    RPL+DiffEEL{\tiny\textcolor{gray}{[Ours]}}&\textbf{71.42}& \textbf{1.28}&\textbf{96.51} & \textbf{0.34} &84.10&12.80 &84.50 &0.80&\textbf{84.31} &3.80  \\
    \hline
    \end{tabular}
  }
  \label{tab:experiment FS_SMIYC_test}
\end{table*}

\subsection{Experimental results on other benchmarks}
\tablename~\ref{tab:experiment FS_SMIYC} presents the comparison results in the validation set of FS-L\&F, FS-Static, SMIYC-Obstacle and RoadAnomaly.
The methods includes Maximum Softmax~\cite{hendrycks2016baseline}, Synboost~\cite{di2021pixel}, SML~\cite{jung2021standardized}, Meta\_OoD~\cite{chan2021entropy}, DenseHybrid~\cite{grcic2022densehybrid}, PEBAL~\cite{tian2022pixel}, RPL~\cite{liu2023residual} and M2A~\cite{rai2024mask2anomaly}, where RPL~\cite{liu2023residual} and M2A~\cite{rai2024mask2anomaly} are previous SOTA methods.
To validate the effectiveness of our proposed DiffEEL, we evaluate the performance of RPL and M2A after combining our DiffEEL.
The results indicate that integrating our proposed DiffEEL achieves substantial improvement in the overall performance.
For the RPL model, we achieve an average performance gain of around 6.90\% in $\rm{AUPRC}$ and 3.58\% in $\rm{FPR}$ by utilizing the proposed diffusion synthesizer and EEL strategy.
When compared with RPL+CoroCL, which incorporates an extra projection module for conservative learning and extra training data from the COCO's inlier objects, our method outperforms it by around 2.94\% in $\rm{AUPRC}$. 
For the M2A model, combining the proposed DiffEEL leads to the greatest improvement in the FS-L\&F, SMIYC-Obstacle and ours ComsAmy benchmark.
Integrating the DiffEEL brings an overall performance gain of around 2.79\% in $\rm{AUPRC}$ and 0.86\% in $\rm{FPR_{95}}$.
The comparison results in the test set of Fishyscapes and SMIYC are in Table~\ref{tab:experiment FS_SMIYC_test}, where all methods are uploaded to the public website and performance is reported by benchmark authorities.
The results indicate that our method is well generalized in all benchmarks and achieves the highest overall $\rm{AUPRC}$ and the second-best $\rm{FPR}_{95}$. 
%


\begin{table*}[t]
\renewcommand\arraystretch{1.1}
  \centering
  \caption{The ablation study. Methods are evaluated on the validation set of FS-L\&F, RoadAnomaly, and ComsAmy benchmarks. The contribution of each proposed component is discussed.}\label{tab:experiment ablation}
  \resizebox{\textwidth}{!}{
  \begin{tabular}{ccc|c|cc|cc|cc|cc}
    \hline
    \multirow{2}*{\thead{Diffusion \\ synthesizer}}&\multirow{2}*{\thead{ EEL\\ anomaly score }}&\multirow{2}*{\thead{EEL \\training loss }}&\multirow{2}*{\thead{Amount of \\ anomalous objects }} & \multicolumn{2}{c|}{FS-L\&F} &\multicolumn{2}{c|}{RoadAnomaly} &\multicolumn{2}{c|}{ComsAmy}&\multicolumn{2}{c}{Overall} \\
    \cline{5-12}
     & & & &\thead{$\rm{AUPRC}~\uparrow$} &$\rm{FPR}_{95}\downarrow$ & \thead{$\rm{AUPRC}~\uparrow$} &$\rm{FPR}_{95}\downarrow$ & \thead{$\rm{AUPRC}~\uparrow$}&$\rm{FPR}_{95}\downarrow$ & \thead{$\tilde{\rm{AUPRC}~\uparrow}$}&$\tilde{\rm{FPR}_{95}\downarrow}$\\  
    \hline
    \xmarkg & \xmarkg &\xmarkg &44,500 &66.73 &3.66 &67.66 &25.73 &59.43 & 50.73 & 53.37 & 17.52 \\    
    \cmark & \xmarkg &\xmarkg &44,500 &70.99 &1.67 &67.67 &24.84 &63.87 &42.99 & 59.50 & 14.51 \\  
    \cmark & \cmark &\xmarkg &44,500 &75.54 &1.56 &68.32 &21.24 &66.01 &39.79 & 62.31 & 12.50 \\  
    \cmark & \cmark &\cmark &44,500 & \textbf{77.87} &\textbf{1.12} &\textbf{74.22} &\textbf{16.74} &\textbf{66.12} &\textbf{38.76} & \textbf{65.43} & \textbf{9.49}\\  
    \hline
    \cmark & \cmark &\cmark &11,125 &72.94 &2.61 &65.73 &19.87 &62.32 &48.81 & 58.33 & 12.61\\  
    \cmark & \cmark &\cmark &22,250 &75.47 &1.68 &69.28 &18.55 &64.67 &42.48 & 61.74 & 11.36\\  
    \cmark & \cmark &\cmark &44,500 &\textbf{77.87} &\textbf{1.12} &\textbf{74.22} &\textbf{16.74} &\textbf{66.12} &\textbf{38.76} & \textbf{65.43} & \textbf{9.49}\\ 
    \hline
  \end{tabular}}
\end{table*}

\subsection{Ablation Study}
In \tablename~\ref{tab:experiment ablation}, we explore the contribution of our proposed diffusion-based data synthesizer, EEL anomaly score, and EEL training loss in the 3 most challenging validation sets: FS-L\&F, RoadAnomaly, and ComsAmy. 
The original RPL model is utilized as our baseline model.
We find that by utilizing the diffusion-based data synthesizer, the model achieves a performance gain of 6.13\% in $\rm{AUPRC}$ and 3.01\% in $\rm{FPR}$ than using copy-paste synthesizer.
By utilizing the EEL anomaly score, the model further improves its performance by 2.81\% in $\rm{AUPRC}$ and 2.01\% in $\rm{FPR}$.
By utilizing the EEL training loss to better distinguish misleading inliers and outliers, the model achieves its best performance by improving 3.12\% in $\rm{AUPRC}$ and 3.01\% in $\rm{FPR}$.  

The training data plays a crucial role in improving the model's performance. 
We also explore the influence of using different sets of anomaly training data that are generated by using different amounts of anomalies.
We test the utilization of 25\%, 50\%, and 100\% of the total amount of anomalous objects to generate the training data. The average $\rm{AUPRC}$ and $\rm{FPR}$ of models trained with data using the aforementioned amount of anomalies are 58.33\% and 12.61\%, 61.64\% and 11.36\%, 65.43\% and 9.49\%, respectively.
The result shows that utilizing more diverse anomalies can improve the model's performance and generalization.
This indicates the great potential of using our diffusion-based anomaly synthesizer to augment the training data for the open-world anomaly segmentation challenge by automatically generating more diverse anomalous objects.

\section{Conclusion and Discussion}
This paper introduces the ComsAmy benchmark, the first anomaly segmentation benchmark designed to comprehensively evaluate anomaly segmentation models in open-set environments.
The ComsAmy benchmark emerges as the most challenging anomaly segmentation benchmark that encompasses adverse weather, intricate driving situations, and a diverse range of anomalies to simulate complex scenarios in open-world situations.
Our extensive evaluation of several state-of-the-art anomalous segmentation models reveals that these methods demonstrate significant deficiencies in such challenging scenarios, highlighting considerable safety risks when deploying such models in real-world applications. 
%

Additionally, this paper proposes a diffusion-based anomaly data synthesizer that automatically generates diverse and high-quality anomalous images, significantly enhancing the training dataset for tackling open-world anomaly segmentation challenges.
By training with more realistic and diverse anomalous images, the model improves its performance in $\rm{AUPRC}$ and $\rm{FPR}$ than using copy-pasting generated images by a great margin. 
Meanwhile, by incorporating a more robust anomaly score that integrates the complementary information from energy and entropy, a novel energy-entropy learning strategy is proposed to improve the model's robustness and generalization under complex and diverse situations in the open-world environment.
Notably, the proposed DiffEEL can be easily plugged into existing methods to greatly enhance their overall performance on anomaly segmentation under all benchmarks.

\bibliographystyle{IEEEtran}
\bibliography{TIP}
\end{document}